\documentclass{poster23}
\usepackage{babel}%
\usepackage{amsmath}
\usepackage{pifont,amssymb}
\usepackage{caption} 
\usepackage{graphicx}
\usepackage{wrapfig}
\usepackage{caption}
\usepackage{subcaption}

\begin{document}

%
\title{mnmDTW: An extension to Dynamic Time Warping for Camera-based Movement Error Localization}
%
\headtitle{S. DILL, M. ROHR, mnmDTW FOR CAMERA-BASED MOVEMENT ERROR LOCALIZATION}

%
\author{Sebastian DILL\affiliationmark{1}, Maurice ROHR\affiliationmark{1}}
%
\affiliation{%
\affiliationmark{1}KIS*MED -- AI Systems in Medicine, Technische Universität Darmstadt, Merckstraße 25, Darmstadt, Germany }
  \email{dill@kismed.tu-darmstadt.de, rohr@kismed.tu-darmstadt.de}

\maketitle


\begin{abstract}
In this approach, we use Computer Vision (CV) methods to extract pose information out of exercise videos. We then employ Dynamic Time Warping (DTW) to calculate the deviation from a gold standard execution of the exercise. Specifically, we calculate the distance between each body part individually to get a more precise measure for exercise accuracy. We can show that exercise mistakes are clearly visible, identifiable and localizable through this metric.
\end{abstract}



\section{Introduction}
Physical therapy is a crucial step in the treatment of many injuries and diseases. One example is hemophilia, where physiotherapy and rehabilitation can help prevent disabilities and preserve a patient's autonomy~\cite{Heijnen}. While ideally, physical therapy is performed under supervision of a medical professional who can offer individual and immediate feedback, most people do not have the resources to visit a training session regularly. Furthermore, home exercises have been shown to be beneficial to the healing process~\cite{proffitt} even without supervision through an expert. On the other hand, wrong executions, misjudgement of one's fitness level, and overexertion might lead to an inefficient training or even worse, serious injuries~\cite{Jones}. To mitigate these problems, an automated evaluation system can be applied to assess the quality of exercise execution and lessen the need for human supervision. Building on the advances in computer vision in recent years, significant research has been conducted on video-based human pose estimation and motion capture. One of the most commonly used tools to extract pose data from videos is MediaPipe Pose based on the BlazePose model~\cite{blazepose}.\\
Dynamic Time Warping (DTW) was established as a fundamental method to estimate the distance between two time-series. While originally conceived for the one-dimensional case, it has also been extended to the multi-dimensional DTW (mDTW)~\cite{mDTW}. As such, it has been applied in the evaluation of human movements, for example in the works of Sempena et al. \cite{Sempena} and Adistambha et al. \cite{Adistambha}. However, their approaches only evaluate the movement as a whole without incorporating the causes of the error. This information is critical when giving the patient feedback on how they should improve their exercises. Liu and Chu~\cite{Liu} address this shortcoming and propose a camera-based machine learning exercise evaluation system that assesses how well an exercise is performed based on posture data extracted from videos, where they not only identified the overall correctness of the exercises but also which body part was responsible for the wrong posture. However, they do not apply DTW to their approach, but instead use domain knowledge to create metrics that they then feed to a deep learning network.\\
In this paper, we present an approach that has both the advantages of the DTW-based solution (low computational complexity, high speed) and the ability to evaluate the error for each body part individually. We propose a new multi-layer normalized multi-dimensional DTW (mnmDTW) approach that is capable of assessing individual body parts and test it on an example exercise. Our evaluation shows very promising results as we can not only correctly classify predefined movement execution errors but also describe them qualitatively in terms of localization and type.

\begin{figure*}[ht!]
    \centering
    \begin{subfigure}{.66\textwidth}
    \centering
    \includegraphics[trim={1mm 1mm 1mm 1mmm},clip,width=\textwidth]{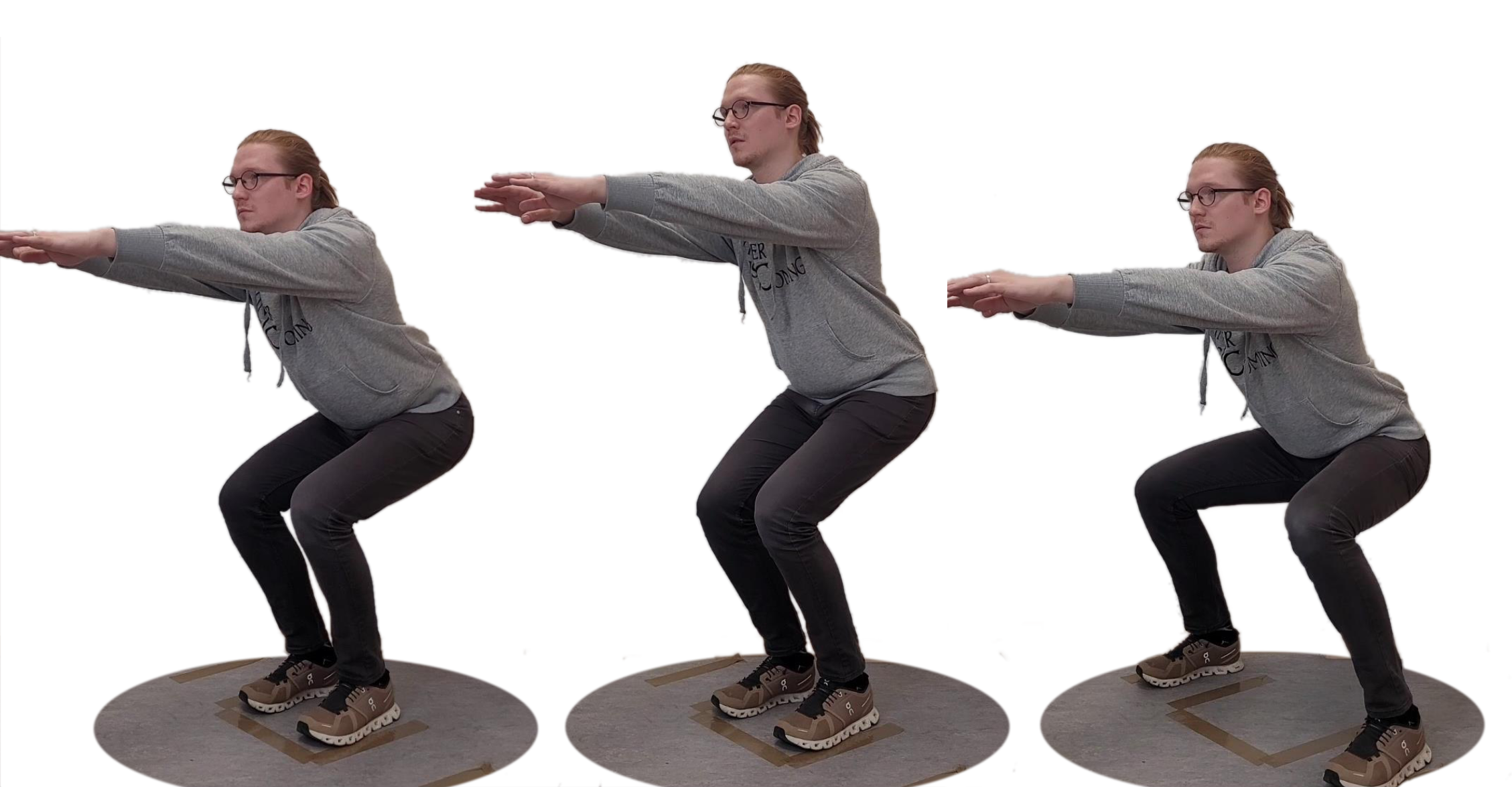}
    \captionsetup{width=0.9\linewidth}
    \caption{From left to right: \textit{Correct} movement; \textit{Mistake 1}: not going low enough and \textit{Mistake 2}: having feet too wide.}
    \label{fig:screenshots_down}
    \end{subfigure}
    \begin{subfigure}{.33\textwidth}
    \centering
    \includegraphics[trim={1mm 1cm 1cm 1.5cm},clip,width=\textwidth]{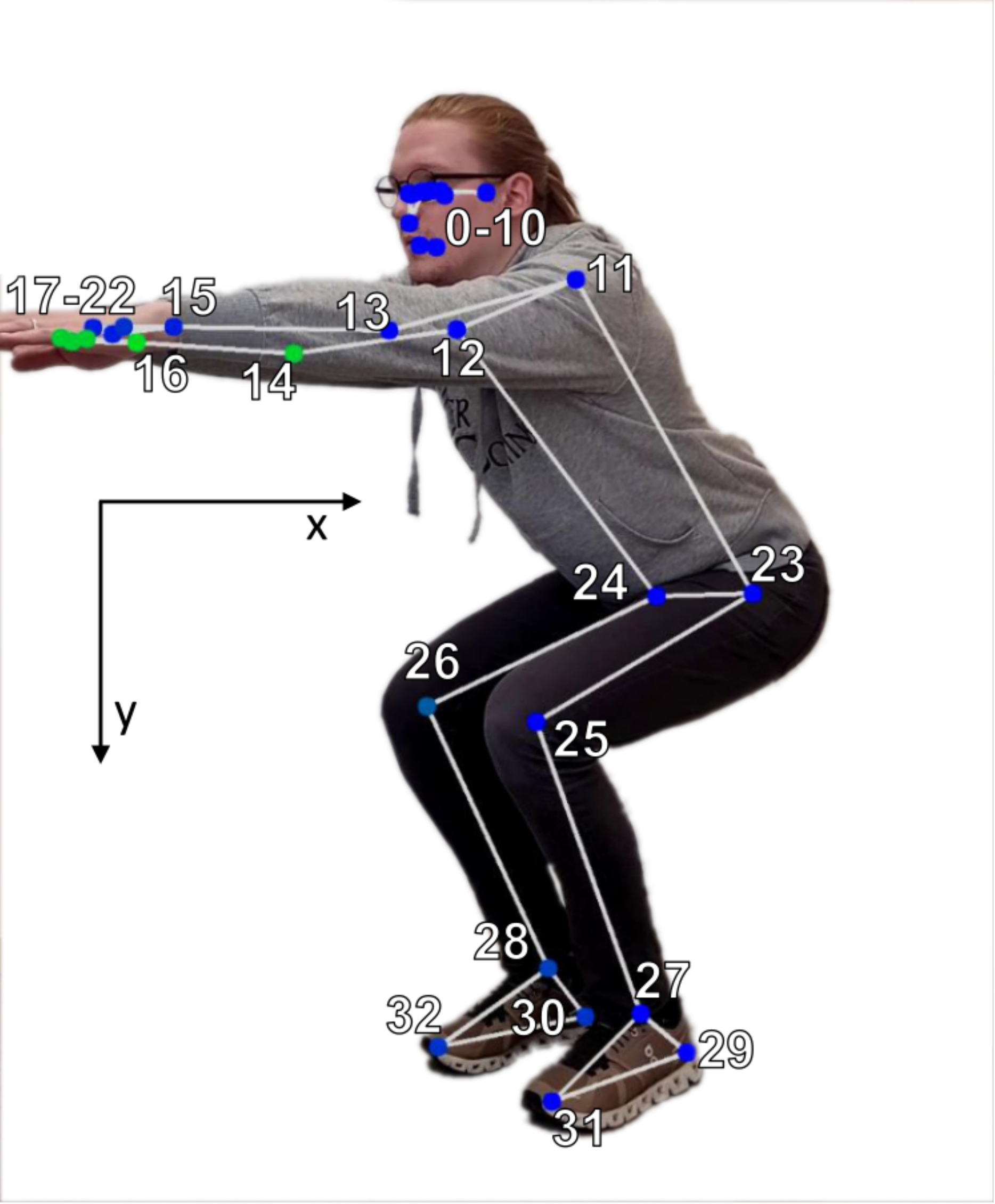}
    \captionsetup{width=0.9\linewidth}
    \caption{visibility is color-coded from blue (1) to green (0).}
    \label{fig:mediapipe}
    \end{subfigure}
    \caption{Visualization of (a) the exercise, where the lowest point of movement is shown for all performed variations and\\ (b) the MediaPipe Pose output, consisting of x-y-camera coordinates of 33 different landmarks.}
\end{figure*}

\section{Experiment}
In this paper, we focused on the squat exercise, since it is a stationary movement that can be done without special equipment and has some defined mistakes. All exercises were recorded by a single camera with $1920\times1080$ pixels and 30 frames per second. We recorded RGB-videos of a single participant performing the exercise 18 times, distinguished into three different categories, depending on the execution quality. Ten executions were considered \emph{correct}, which we defined by the participant's feet being about as wide as their shoulders and the minimum knee angle being close to $90^{\circ}$. Two common mistakes are also considered and recorded four times each. \textit{Mistake 1} is defined by the participant not going low enough and their knee angle staying well above $90^{\circ}$. \textit{Mistake 2} is defined by the participant's feet being further apart than shoulder width. Figure \ref{fig:screenshots_down} shows the lowest point for one execution each of all three variations. From the ten videos showing the \textit{correct} exercise, one was randomly selected as the ``gold standard''. The remaining 17 videos are considered test videos.

\section{Methods}
\begin{figure*}[ht!]
\centering
\includegraphics[trim={0mm 11cm 1cm 0cm},clip,width=\textwidth]{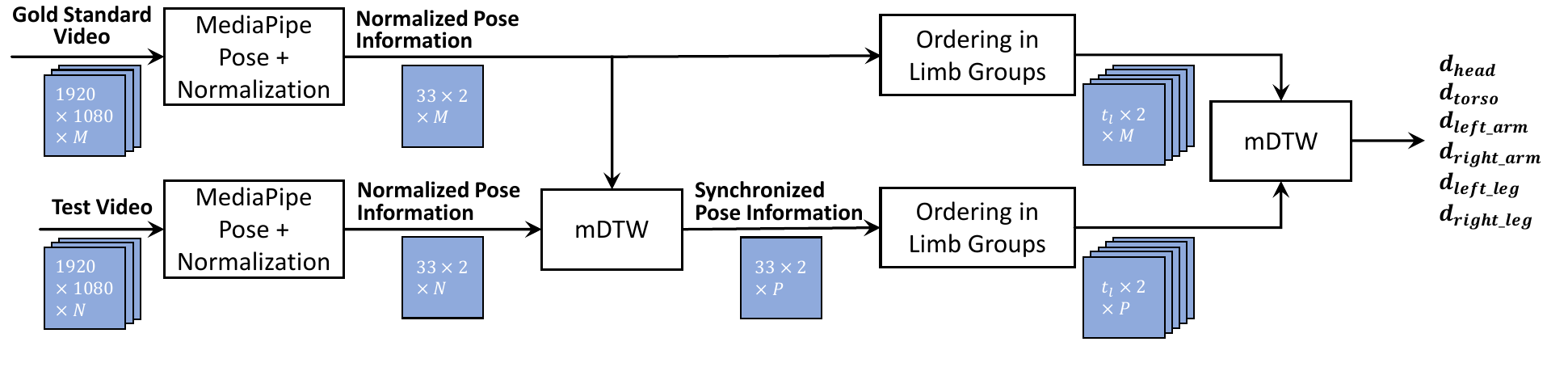}
\captionsetup{width=\linewidth}
\caption{The general idea behind mnmDTW for movement error localization. After extracting and normalizing pose data from videos, a first mDTW is done to synchronize the test video data of length $N$ with the predefined gold standard recording of length $M$ to receive a time signal of length $P=\text{max}(M,N)$. Then, the synchronized data is ordered in $L$ limb groups of dimension $t_l$ and a second mDTW is performed for each group to receive group-specific distance metrics.
} 
\label{fig:flowchart}
\end{figure*}

For each test video, joint positions are extracted, normalized and synchronised to the gold standard with a first mDTW. After combining the joints to groups representing a limb each, a second mDTW is calculated to receive group-specific mnmDTW values. The overall process can be seen in Fig. \ref{fig:flowchart} and is explained in the following chapter.

\subsection{Extraction of Pose Information}
All 18 videos are first processed with the MediaPipe Pose library. The MediaPipe output consists of x-y-z-camera-coordinates of 33 different pose landmarks as well as an estimate for visibility between 0 and 1 for every landmark. For this work, we only used the x-y-camera-coordinates. Since these are given in pixels, with the origin being in the upper left corner of the image, we need to normalize the coordinates with z-normalization to remove the bias. For an overview over the provided landmarks, see Fig. \ref{fig:mediapipe}. 

\subsection{Dynamic Time Warping}
The general idea of DTW is to measure similarity between two time-series $\boldsymbol{x}=[x_1,\dots,x_i,\dots,x_M],~\boldsymbol{y}=[y_1,\dots,y_j,\dots,y_N]$, that can have different speeds or lengths. DTW is a non-linear algorithm that disregards the exact timestamps at which observations occur. Instead, it finds the optimal ordering of timestamps by minimizing the Euclidean distance between the series under all admissible temporal alignments. Each alignment is characterized by its alignment path $\pi = [(i_1,j_1),\dots,(i_p,j_p),\dots,(i_{P},j_{P})],~P=\text{max}(M,N)$ mapping indices from one series to the other. For an alignment to be admissible, its path has to fulfill these constraints:
\begin{enumerate}
    \item Each point from one series must be matched to at least one point from the other, in a monotonically increasing way: $i_{p-1}\leq i_p\leq i_{p-1}+1$ and $j_{p-1}\leq j_p\leq j_{p-1}+1$.
    \item The first index from one series must be matched to the first index of the other series: $\pi_1=(1,1)$. The same applies to the last indices: $\pi_{P}=(M,N)$.
\end{enumerate}
All mapping paths $\pi$ that satisfy these constraints span a set of possible paths $\mathcal{A}$. The overall DTW error is then calculated as 
\begin{equation}
    d_{\text{DTW}}(\boldsymbol{x},\boldsymbol{y}) = \text{min}_{\pi\in\mathcal{A}(\boldsymbol{x},\boldsymbol{y})}\sqrt{\sum_{(i,j)\in\pi}d_{ij}^2},
    \label{eq:DTW}
\end{equation}
with the Euclidean distance $d_{ij} = \sqrt{(x_i-y_j)^2}$.\\
In the case of K-dimensional time series, $\boldsymbol{x}$ and $\boldsymbol{y}$ become matrices $\boldsymbol{X} = [\boldsymbol x_1,\dots, \boldsymbol x_i,\dots,\boldsymbol x_M]$, with $\boldsymbol{x}_i = [x_{i,1},\dots,x_{i,k},\dots,x_{i,K}]$ and $\boldsymbol{Y} = [\boldsymbol y_1,\dots, \boldsymbol y_j,\dots,\boldsymbol y_N]$, with $\boldsymbol{y}_j = [y_{j,1},\dots,y_{j,k},\dots,y_{j,K}]$. The distance $d_{ij}$ then can be calculated as
\begin{equation}
    d_{ij} = \sqrt{\sum_k (x_{i,k}-y_{j,k})^2}.
    \label{eq:eucl_distance}
\end{equation}

\subsection{Multi-layer Normalized Multi-Dimensional Dynamic Time Warping}
We calculate the multi-dimensional DTW (mDTW) as defined in equations \ref{eq:DTW} and \ref{eq:eucl_distance} over all 66 dimensions (x-y-coordinates for 33 landmarks) to align all test recordings to the gold standard. To evaluate which body part contributes the most to the error, we then combine the landmarks into limb groups. The mapping of landmarks to limb group can be seen in Table \ref{tab:limb_grouping}. Then, the mDTW distance between the gold standard and the aligned recordings are calculated separately both for each limb group and for x- and y-coordinates. The resulting mnmDTW values are a metric for how similar specific body parts move in comparison to the gold standard. Furthermore, the separation into x- and y-coordinates gives us more information on the type of error.

\begin{table}[ht!]
    \centering
    \begin{tabular}{|c|c|c|}
        \hline
         \textbf{landmark} &  \textbf{limb group} & dimension $t_l$\\
         \hline
         0-10 & head & 11\\
         11, 12, 23, 24 & torso & 4\\
         13, 15, 17, 19, 21 & left\_arm & 5\\
         14, 16, 18, 20, 22 & right\_arm & 5\\
         25, 27, 29, 31 & left\_leg & 4\\
         26, 28, 30, 32 & right\_leg & 4\\
         \hline
    \end{tabular}
    \caption{Mapping of landmarks (see fig.~\ref{fig:mediapipe}) to limb group.}
    \label{tab:limb_grouping}
\end{table}
For better comparability and interpretability, another five \textit{correct} exercises were selected as a control group. The corresponding videos were used to calculate baseline mnmDTW values for each limb that were then averaged. All other mnmDTW distances were normalized by dividing by the baseline average value of the corresponding limb. This makes the mnmDTW values more intuitive. Generally speaking, all values of 1 and below are regarded as \emph{good}. On the other hand, the less the movement matches the gold standard the greater the mnmDTW value.

\section{Results}
\begin{figure*}[ht]
    \centering
    \begin{subfigure}{.33\textwidth}
    \centering
    \includegraphics[trim={.5cm 0.5cm 2cm 1cm},clip,width=\columnwidth]{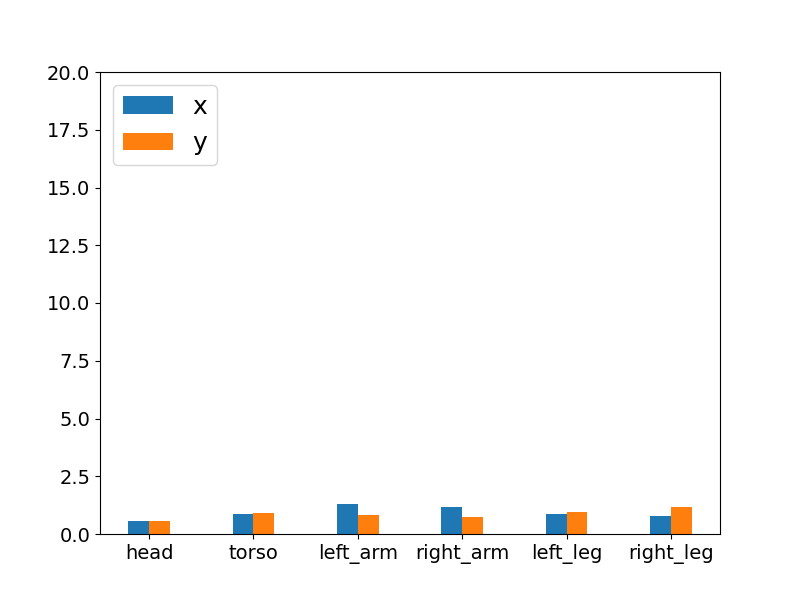}
    \caption{\textit{correct}}
    \label{fig:res1}
    \end{subfigure}
    \begin{subfigure}{.33\textwidth}
    \centering
    \includegraphics[trim={.5cm 0.5cm 2cm 1cm},clip,width=\columnwidth]{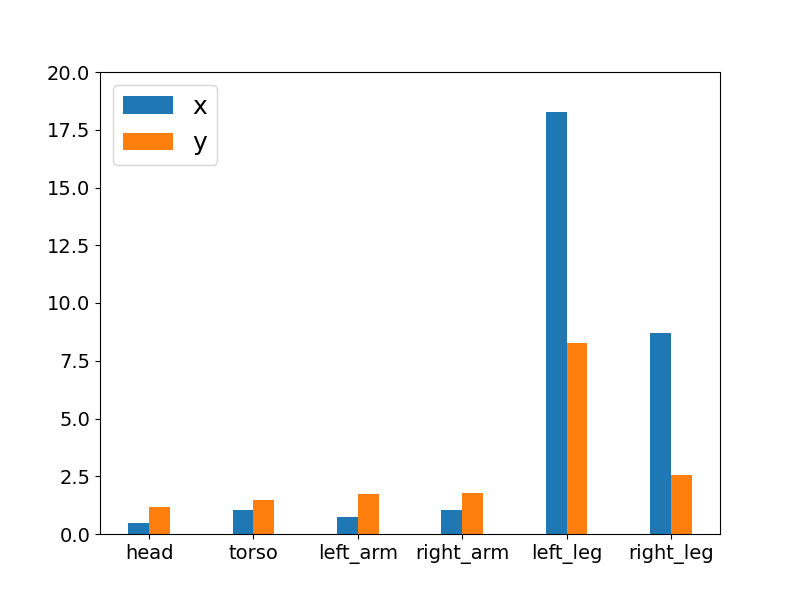}
    \caption{\textit{Mistake 1}: having your feet too wide.}
    \label{fig:res2}
    \end{subfigure}
    \begin{subfigure}{.33\textwidth}
    \centering
    \includegraphics[trim={.5cm 0.5cm 2cm 1cm},clip,width=\columnwidth]{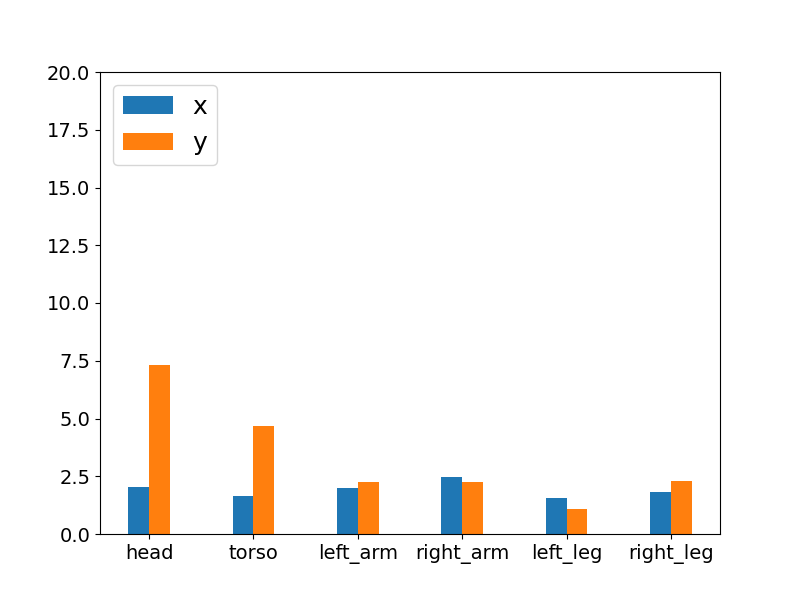}
    \caption{\textit{Mistake 2}: not going deep enough.}
    \label{fig:res3}
    \end{subfigure}
    \caption{mnmDTW values for three example exercises.}
\end{figure*}

Figures \ref{fig:res1},~\ref{fig:res2} and \ref{fig:res3} show mnmDTW values for three representative example exercises, one from each class. As expected, for the \emph{correct} exercise, the values are approximately one for all limb groups. 
For \textit{mistake 1}, ``having your feet too wide'', the values for the limb groups ``left\_leg'' and ``right\_leg'' are far higher, especially in the x-axis, while the values corresponding to the upper body parts remain close to one. Since this mistake is defined solely by a horizontal offset of the feet, this observation matches our expectations. It can also be seen that the mnmDTW value is even higher for the left leg. This can be explained through the the projection of the movement into camera-coordinates. As can be seen from Fig. \ref{fig:screenshots_down}, the right leg is further away from the camera and therefore, the displacement is smaller in camera-coordinates. The projection also explains why the left leg has a high metric for the y-coordinate. 
For mistake 2, ``not going deep enough'', all limbs have an increased mnmDTW value, mainly in the y-axis. Furthermore, it can be observed that the values are increasing with the limb's height and the maximum error is achieved for the head. Again, this matches the human observations when looking at Fig.~\ref{fig:screenshots_down}: Since the error is caused by not bending the knees far enough, there is little difference in the positions of the feet and lower legs. The further up the limbs are, the higher the offset gets, with especially high errors for arms and head.

\section{Discussion}
For all of our recordings, the mnmDTW values accurately describe the visible deviation from the gold standard movement. This not only enables a classification based on predefined mistakes, but also, our results indicate the method's potential to qualitatively and quantitatively evaluate and describe movements. It is possible not only to tell which limbs are responsible for the error, but also, by looking at the coordinates separately, to estimate where the error might be located and what it might look like. This way, even unknown movement errors could be identified and corrected.

\section{Conclusion and Outlook}
This small-scale test shows very promising results for the mnmDTW approach. As discussed, the metric not only allows for simple classifications into predefined classes of movement quality, but also enables us to make qualitative and quantitative statements about what the cause for low exercise quality might be. However, our experiment was very limited in size and variety. Overall, only 18 recordings were made of a single person from only one camera angle. A larger experiment would be the logical next step. Here, the influence of different looks, camera angles, fitness levels and exercises can be researched. There are also several aspects how the mnmDTW metric could be improved in the future. First, the metric could be extended to 3D-coordinates, which would reduce the expected dependency on the camera properties and recording situation. Also, the degree to which the coordinates are grouped into limbs, is a parameter of interest. Making the groups smaller enables even more precise localization of the error, but is also expected to introduce more noise to the metric.

\section*{Acknowledgements}
The research described in the paper was supervised by Prof. C. Hoog Antink, TU Darmstadt.

\setlength{\intextsep}{0pt}%
\setlength{\columnsep}{2pt}%
\begin{authorcv}{}

\begin{wrapfigure}{l}{0.15\textwidth}\centering
\includegraphics[trim={0.5cm 1cm 0 0.5cm},clip,width=0.15\textwidth]{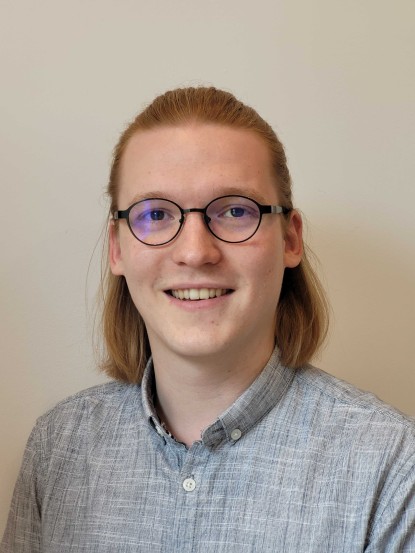}
\end{wrapfigure} 
\textbf{Sebastian DILL} was born in München, Germany, in 1996, finished his Abitur in 2014 and then began his studies in Darmstadt. In 2021 he finished his master degree in electrical engineering at the Technical University Darmstadt where he is currently working towards a Ph.D. degree in electrical engineering at the AI Systems in Medicine group. His research interests include non-obtrusive pose estimation, movement analysis and physical therapy.

\begin{wrapfigure}{l}{0.15\textwidth}\centering
\includegraphics[trim={2.5cm 0 2.5cm 0},clip,width=0.15\textwidth]{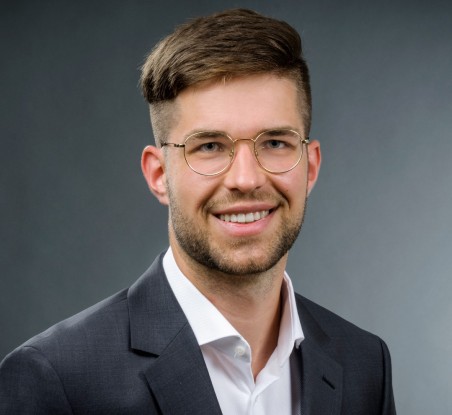}
\end{wrapfigure} 
\textbf{Maurice ROHR} was born in Mannheim, Germany, in 1995, finished his Abitur in 2014 and then began his studies in Darmstadt. In 2020 he finished his master degree in electrical engineering at the Technical University Darmstadt where he is currently working towards a Ph.D. degree in electrical engineering at the AI Systems in Medicine group. His research interests include medical sensor fusion, imaging technologies and simulation.

\end{authorcv}

\end{document}